\documentclass{article}

\usepackage{arxiv}
\usepackage[utf8]{inputenc}
\usepackage[T1]{fontenc}
\usepackage{url}
\usepackage{booktabs}
\usepackage{amsfonts}
\usepackage{nicefrac}
\usepackage{microtype}
\usepackage{lipsum}
\usepackage{graphicx}
\usepackage{subfig}
\usepackage{float}
\usepackage{longtable}
\usepackage{multirow}
\usepackage{listings}

\usepackage{listings}
\usepackage{xcolor}

\definecolor{codegreen}{rgb}{0,0.6,0}
\definecolor{codegray}{rgb}{0.5,0.5,0.5}
\definecolor{codepurple}{rgb}{0.58,0,0.82}
\definecolor{backcolour}{rgb}{1,1,1}

\lstdefinestyle{mystyle}{
    backgroundcolor=\color{backcolour},   
    commentstyle=\color{codegreen},
    keywordstyle=\color{orange},
    numberstyle=\tiny\color{codegray},
    stringstyle=\color{darkgray},
    basicstyle=\ttfamily\footnotesize,
    breakatwhitespace=false,         
    breaklines=true,                 
    captionpos=b,                    
    keepspaces=true,                 
    numbers=left,                    
    numbersep=5pt,                  
    showspaces=false,                
    showstringspaces=false,
    showtabs=false,                  
    tabsize=2
}

\title{Detection of news written by the ChatGPT through authorship attribution performed by a Bidirectional LSTM model}

\author{
  Amanda Ferrari Iaquinta \\
  Department of Electrical Engineering\\
  IFSP - Federal Institute of Education, Science and Technology of São Paulo\\
  Piracicaba - SP - Brazil \\
  \texttt{a.ferrariiaquinta@gmail.com } \\
  \and
  Gustavo Voltani von Atzingen \\
  IFSP - Federal Institute of Education, Science and Technology of São Paulo\\
  Rua, Av. Diácono Jair de Oliveira, 1005 \\
  Piracicaba - SP - Brazil \\
  \texttt{gustavo.von@ifsp.edu.br} \\ 
}

\begin{document}
\maketitle

\begin{abstract}
    The large language based-model chatbot ChatGPT gained a lot of popularity since its launch and has been used in a wide range of situations. This research centers around a particular situation, when the ChatGPT is used to produce news that will be consumed by the population, causing the facilitation in the production of fake news, spread of misinformation and lack of trust in news sources. Aware of these problems, this research  aims to build an artificial intelligence model capable of performing authorship attribution on news articles, identifying the ones written by the ChatGPT. To achieve this goal, a dataset containing equal amounts of human and ChatGPT written news was assembled and different natural processing language techniques were used to extract features from it that were used to train, validate and test three models built with different techniques. The best performance was produced by the Bidirectional Long Short Term Memory (LSTM) Neural Network model, achiving 91.57\% accuracy when tested against the data from the testing set.
\end{abstract}


\section{Introduction}
Large Language Models (LLMs) are a category of deep learning algorithms trained to understand and generate human language \cite{chang2023survey}. Among other areas, these algorithms show high efficiency when applied to problems involving language processing due to the fact that they use a specific neural network architecture created in 2017, called Transformers \cite{vaswani2017attention}. Since its creation, this architecture has enabled great advances in natural language processing (NLP) studies and techniques, but it was from the launch of the large language based-model chatbot ChatGPT, in November 2022, that it gained prominence. This fact can be visualized on Figure 1 that shows the worldwide Google search interest \cite{googleTrends} relative to the term ‘Large Language Models’ between January 2017 and September 2023. The y-axis represents the interest relative to the highest point on the chart, with a value of 100 representing the peak popularity of the term. Compared to previous months, it is possible to note a rapid growth in interest from November 2022 until September 2023, which is the month this article was written.

\begin{figure}
  \centering
  \includegraphics[width=0.98\linewidth, height=9cm]{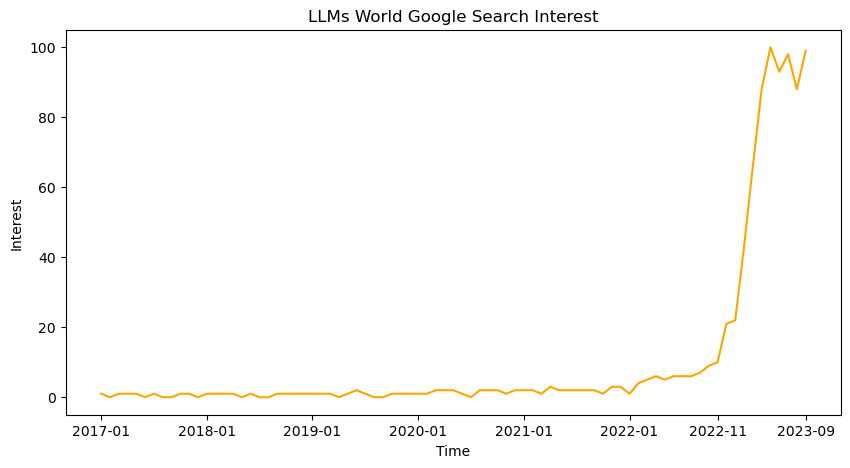}
  \caption{LLMs world Google search interest.}
  \label{fig:trends}
\end{figure}

The mentioned chatbot is a natural language processing tool based on the transformer GPT (Generative Pretrained Transformer), trained to have human-like conversations and provide text solutions for the most diverse questions and requests \cite{bird2009natural}, being able to write human-like texts, essays, news articles, documents, e-mails, fragments of code for computers and many other textual constructions. As a result of its training, this tool can be used in a wide range of situations \cite{kasneci2023chatgpt}, \cite{juola2008authorship}, \cite{deng2022benefits} causing positive and negative impacts.

This research centers around a particular situation, when the ChatGPT is used to produce news that will be consumed by the population, causing two main problems. The first problem is the facilitation in the production of fake news due to the fact that the generated text is guided and based on the request provided by the user that may or may not contain false information that, until the date of this study, will not be fact checked by the chatbot before writing a news about it. The second problem is a direct result from the first, consisting of the fact that people end up losing confidence in the information they read, not knowing which source to trust, leading to chaos and misinformation \cite{wach2023dark}.

The mentioned problems are reinforced by the result of a study carried out by this research and explained in the following sections, which showed that the degree of sophistication of the textual productions generated by the chatbot is so high that it is extremely difficult for the human being to differentiate whether or not a text was written by the ChatGPT. Concerned with the exposed situation, this research uses machine learning and deep learning techniques to perform authorship attribution on news and identify when a news article was written by the ChatGPT. 

The authorship attribution (AA) task consists of correctly assigning an author, from a specific set, to a text of unknown authorship, given sample texts produced by the possible authors \cite{singh2021authorship}. Successfully completing this task is of extreme importance in many applications such as plagiarism detection, forensic investigations and literature and history studies \cite{sari2018topic}, \cite{juola2008authorship}. The general approach used to execute this task is by extracting different stylometric features from the sample texts and using some classifier algorithm to recognize patterns and make authorship attributions based on the extracted features \cite{singh2021authorship}. 

Stylometric features can be defined as information contained in a text that can be used to characterize and differentiate its author. As studied by \cite{stamatatos2009survey}, these features can be classified in five groups: lexical, character, syntactic, semantic and application-specific. Concisely, the lexical group views a text as a sequence of tokens and the only requirement to extract this type of feature is a previous text tokenization, while the character group views a text as a sequence of characters and the features are extracted from a character level. The syntactic group analyzes the syntactic structures that form a text, relying on the tendency of authors to use similar syntactic patterns, and the semantic group consists of features that consider the context, emotions, and sentiments present in a text. Lastly the application-specific group consists of features that can only be extracted from texts that have specific contents. 

Many different classifier algorithms were used in the literature to learn patterns present in extracted features and perform AA. \cite{shrestha2017convolutional} presented a Convolutional Neural Network (CNN) model that uses a sequence of character n-grams as input for AA of tweets using a dataset with approximately 9.000 Twitter users with up to 1.000 tweets each; \cite{hitschler2017authorship} also proposed a CNN approach but used different feature extraction processes to perform AA in a dataset containing articles from 880 different authors divided in segments of 1.500 words; \cite{gomez2018stylometry} used Logistic Regression and Support Vector Machine (SVM) techniques to identify changes in the writing style of 7 authors of novels; \cite{qian2017deep} used Gated Recurrent Unit (GRU) network and Long Short Term Memory (LSTM) network to perform AA using the Reuters 50 50 news dataset and Gutenberg story dataset; \cite{oliva2022improving} tested different LSTM architectures to perform AA on tweets and \cite{jafariakinabad2019syntactic} proposed a Syntactic Recurrent Neural Network model using both CNN and LSTM techniques to perform AA using a closed dataset containing novels from 14 different authors. Different from the previous studies and following the same line of research of this paper, \cite{guo2023close} tested two deep classifiers and a logistic regression model to differentiate answers written by the ChatGPT from the ones written by humans and \cite{pegoraro2023chatgpt} made an assessment of the most recent techniques in ChatGPT writing detection.

In view of the above, this research aims to build an artificial intelligence classification model capable of performing authorship attribution on news articles, identifying the ones written by the ChatGPT. To achieve this goal, different machine learning and deep learning algorithms were used along with various stylometric feature extraction processes.

\section{Methodology}
A dataset containing equal amounts of human and ChatGPT written news was assembled through news gathering of online news platforms and news generation with the ChatGPT API. The collected news were subjected to a series of different natural language processing (NLP) operations aiming to produce meaningful data that was used to train, validate and test the artificial intelligence classifiers proposed in this paper. Three classification models were built using different techniques and compared in order to find the one that best suited the problem introduced above. The code produced to execute the process and models described in this paper is available on a public repository \cite{amanda_ferrari_iaquinta_2023_8332854} on GitHub.

The dataset assembly process was performed in three steps. In the first step, three well known and reliable news sites were selected: BBC News, TechCrunch and The Verge. For each site, data scraping was performed using a series of Python libraries to obtain the human written news part of the dataset. Three rounds of news scraping were performed, collecting news written in English and published in the periods of January 2022 to June 2022 or January 2023 to April 2023. After each round, the collected news were reviewed by humans to ensure that the content was correctly scraped and did not contain external text such as advertisements and readers comments. The scrapping process was set to search for technology themed news to avoid that the theme became a bias later on the analysis. A total of 623 news articles were collected, 203 from BBC, 252 from Tech Crunch and 168 from The Verge.

In the second step, the scraped news articles along with the OpenAI ChatGPT 3.5 API were used to obtain the ChatGPT part of the dataset. Two interactions with the ChatGPT API occurred for each scrapped news article. First, a summary was requested for each news article. The returned summary was then used in the second interaction requesting a news article based on that summary. The API ‘memory’ feature was not used, this way the news article produced by the ChatGPT was based only in the given summary, having no relation with the news article present in the previous summary request. After this process, each human written news article had a ChatGPT written version. 

The Python code snippet below shows the exact queries used to interact with the ChatGPT 3.5 API. Line 2 contains the query used in the first interaction and line 4 contains the query used in the second interaction. The variables 'text' and 'text\_tokens' present in the code contain a collected news article written by humans and its tokenized version respectively and variable 'summary' contains the news article summary generated in the first interaction. To prevent that the news size became a bias on the classification process, the second interaction limited the ChatGPT news version to have a size in words within a range of plus 50 words or minus 50 words from the size of the human written news version, as shown in the code.

\begin{lstlisting}[language=python]

summary_query = "Summarize in details the following news article. News: " + text

news_article_query = "Write a news story based on the following summary containing a maximum of " + str(len(text_tokens) + 50) + " words and a minimum of " + str(len(text_tokens) - 50) + " words and entitled: " + title + ". Summary: " + summary
\end{lstlisting}

The third and final step was dividing the dataset in three subsets named train, validation and test. The train subset contains 70\% of the dataset data with 870 news and the validation and test subsets each have 15\% of the dataset data with 186 and 190 news respectively. The separation was performed randomly, not implying that the human and ChatGPT versions of a news article stayed in the same set. The violin plot of Figure 2 depicts the news articles size distribution for each data subset. For this depiction, the news articles were tokenized using the NLTK tokenizer and the number of tokens of each news article was considered as its size. Respectively the train, validation and test subsets have medians of 613.5, 601 and 626.5 tokens.

\begin{figure}
  \centering
  \includegraphics[width=0.98\linewidth, height=9cm]{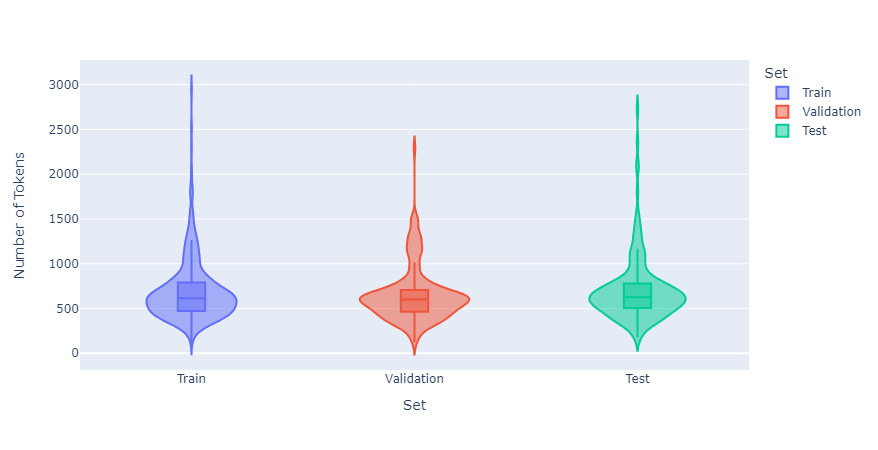}
  \caption{News articles size distribution.}
  \label{fig:stats}
\end{figure}

Once the dataset was assembled and divided, two different data treatments were performed. Both treatments described below used the part-of-speech (POS) tagging technique that is a widely used process for feature extraction on AA problems \cite{hitschler2017authorship}, \cite{jafariakinabad2019syntactic}, \cite{martinez2012part}. This process consists of assigning a grammatical category to a given word using contextual information to make the correct classification. Each category is represented by a tag that is formed by one or more capital characters and the number of possible categories varies according to the chosen classifier and the degree of specification intended. For this research, the NLTK library \cite{bird2009natural} POS tag classifier was used, having 226 possible categories.

Treatment number 1 produced the data used to train, validate and test the first two proposed models that will be described below, and consisted in extracting lexical and syntactic features from the news articles. After studying and evaluating many stylometric features present and not present in the literature \cite{singh2021authorship}, \cite{sari2018topic}, \cite{ramnial2016authorship}, \cite{bhargava2013stylometric}, \cite{bozkurt}, 13 features were extracted from every news article and are described in Table 1. Initially, the news articles were tokenized using NLTK library tokenizer and the first 4 features present on Table 1, classified as statisticals, were extracted directly from the raw tokenized text. It is important to address that for the extraction of the statistical feature number 2 from Table 1, named ‘Stopwords ratio’, the english stopwords set from the NLTK python library was used to identify this type of word, that consists of the most frequent words used in a given language. Following the tokenization, the part-of-speech (POS) tagging process was performed on the tokenized data and after this process each news article no longer corresponded to tokenized raw text but to a sequence of POS tags from which 9 features were extracted and are also described in Table 1. Upon completion of this treatment, each news article was represented by a one dimensional array of 13 numeric features, each one being the value of a feature presented in Table 1, as exemplified bellow:

\begin{eqnarray*}
    ndarray([0.44308452, 0.34832337, 0.1956427, 0.53335732, 0.61453386, 0.18496241, \\
    0.24151003, 0.47051729, 0.63403149, 0.48049792, 0.5156211, 0.11363636, 0.5])
\end{eqnarray*}

Treatment number 2 produced the data used to train, validate and test the third proposed model that will be described below. Initially, the data was tokenized and POS tagged in the same way as described in treatment number 1. Each POS tag belonging to a sequence was then converted to an integer so that each tag sequence corresponded to a one dimension array of integers that would then go through a sequence filling process. In this process, all arrays were brought to the same size of 200 by, if necessary, filling them with zeros to the left until the desired size was reached. Once performed the steps described, each news article was represented by a one dimensional array of 200 integer numbers as exemplified below:

\begin{eqnarray*}
   ndarray([3, 4, 9, ..., 9, 12])
\end{eqnarray*}

\begin{longtable}{|c|p{.6\linewidth}|p{.5\linewidth}|}
  \caption{ Features extracted with treatment number 1.} \\ 
  \hline
  \textbf{Feature}  & \multicolumn{1}{c|}{\textbf{Description}}
  \endhead
  \hline
  1. Ponctuation ratio & Quantity of punctuation tokens divided by total number of tokens. All characters present in the following set were considered punctuation: !"\#\$\%\&\'()*+,-./:;<=>?@[\textbackslash]\^\_\`\{|\}$\sim$ \\
  \hline
  2. Stopwords ratio & Quantity of tokens that correspond to stopwords divided by total number of tokens. \\
  \hline
  3. Average number of tokens per sentence & Number of tokens divided by the total number of sentences. \\
  \hline
  4. Average token length & Number of characters divided by the total number of tokens.\\
  \hline
  5-1. Adjective ratio & Number of tags indicating adjective  divided by the total number of POS tags.\\
  \hline
  6-2. Nouns ratio & Number of POS tags indicating nouns divided by the total number of POS tags.\\
  \hline
  7-3. Conjunction ratio & Number of tags indicating conjunctions divided by the total number of POS tags.\\
  \hline
  8-4. Preposition ratio & Number of tags indicating prepositions divided by the total number of POS tags.\\
  \hline
  9-5. Adverbs ratio & Number of tags indicating adverbs  divided by the total number of POS tags. \\
  \hline
  10-6. WH ratio & Number of tags indicating WH tokens (what, which, what’s, whose …) divided by the total number of POS tags. \\
  \hline
  11-7. Modals ratio & Number of tags indicating modals  divided by the total number of POS tags. \\
  \hline
  12-8. Determiners ratio & Number of tags indicating determiners divided by the total number of POS tags. \\
  \hline
  13-9. Verbs ratio & Number of tags indicating verbs divided by the total number of POS tags. \\
  \hline
\end{longtable}

\section{Results}
Following the increasing order of complexity, the built models consisted in a Extreme Gradient Boosting (XGBoost) classifier, an Artificial Neural Network (ANN) and a Bidirectional Long Short Term Memory (LSTM) Neural Network. Of the three mentioned techniques, the Bidirectional LSTM had the best performance, followed by the ANN and the XGBoost in this order.

The first tested model was the one built with the XGBoost technique, which consists in the application of an ensemble machine learning algorithm based on decision trees that uses gradient boosting \cite{chen2016xgboost}. This algorithm has several parameters that can be adjusted to better perform in different applications. The model used in this research mainly worked with three parameters, setting the booster mode to the gbtree option, the learning rate to 0.02 and the objective to the option binary:logistic. After being trained using the mentioned parameters configuration and the training set outputted from treatment number 1, the model had 81.05\% of accuracy when tested against the data from the testing set. 

The Artificial Neural Network model was tested next with the same training set and had 83.15\% of accuracy when tested against the testing set. The net architecture consisted of 2 hidden dense layers, configured to use the ReLU activation function and to have 16 and 8 units respectivly, followed by a dense layer with only one unit that was configured to use the sigmoid activation function. More complex and bigger architectures were also tested but as they did not produce better results than the one described, the simplest architecture was chosen. 

Analyzing the results obtained with the two models described above and aiming to obtain even better results, a third model was built with a different technique. The chosen technique was the LSTM, a type of Recurrent Neural Network (RNN) that is able to learn from sequential or temporal data, maintaining memory of long and short term dependencies \cite{huang2015bidirectional}. Thus, going through a sequence, the analysis of each term is influenced by features and patterns learned from the past, that is, from the terms that came previous to the current analyzed term. In the context of this research, a news article resulting from treatment number 2 described above is a sequence of POS tags and for each term, that is, POS tag in the sequence, the previous and next terms are known. This fact allowed the use of a Bidirectional LSTM, that differs from a normal LSTM by allowing the previous and the next term to influence the analysis of a given current term. 

The architecture of the built Bidirectional LSTM model is represented in Figure 3 and consisted of a embedding layer, responsible for converting each POS tag from the training set sequences in vectors of size 150, followed by a Bidirectional LSTM layer with 64 units and 2 dense layers set to use ReLu and Sigmoid activations functions respectively with 32 units each. The Bidirectional LSTM layer was configured to have a dropout of 0.2 and a recurrent dropout of 0.2. After training with the training set produced from treatment number 2, this model achieved 91.57\% accuracy when tested against the data from the testing set.

\begin{figure}[H]
  \centering
  \includegraphics[width=0.45\linewidth, height=9cm]{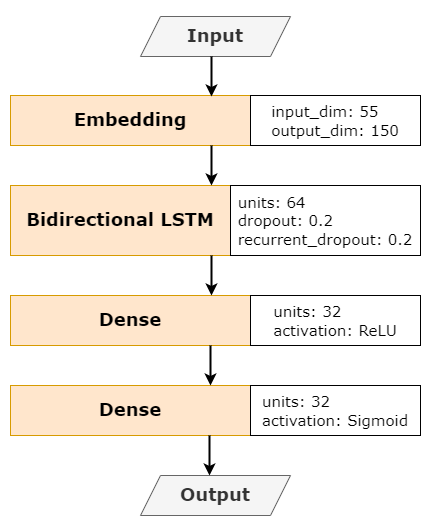}
  \caption{Proposed bidirectional LSTM model.}
  \label{fig:lstm}
\end{figure}

Table 2 presents a comparison between the performances of the three models tested and humans performing the same task proposed for the models. The humans accuracy was calculated using data collected in a study conducted by this research. In this study, each participant was asked to read,  through a website,  5 news articles randomly chosen from the test set used to test the models and classify according to their perception if the news was authored by a human or by the ChatGPT. A total of 63 people completed the study and their average accuracy was 57.78\%.

\begin{table}
\centering
\caption{Performance comparison.}
\label{results}
\begin{tabular}{|c|c|c|}
    \hline \multicolumn{2}{|l|}{} & {Accuracy} \\
    \hline \multirow{3}{*}{Trained Models} & Bidirectional LSTM & 91.57\% \\
    \cline{2-3} & ANN & 83.15\% \\
    \cline{2-3} & XGBoost & 81.05\% \\
    \hline
    Conducted Study & Humans & 57.78\% \\
    \hline
\end{tabular}
\end{table}

\section{Conclusion}
The analysis of the results present on Table 2 shows that the proposed LSTM model performed very well on the assigned authorship attribution task, especially when compared to humans performing the same task, fact that highlights its importance as a tool to help combat the problems caused by the ChatGPT written news that were mentioned above. Although 91.57\% represents a very good accuracy score, it can be improved using a bigger dataset to train, validate and test the model. It is important to mention that this model was not tested on news originated from news sites different from the 3 that were initially selected to compose the dataset, therefore further research may focus on generalizing this approach and make the model effective for any given news source.

\bibliographystyle{unsrt}
\bibliography{references} 

\end{document}